\documentclass[letterpaper, 10 pt, conference]{ieeeconf}  

\IEEEoverridecommandlockouts

\usepackage{algorithm}
\usepackage{algorithmic}
\usepackage{url}
\usepackage{amsmath}
\usepackage{cite}
\usepackage[dvips]{graphicx}
\usepackage{epsfig}
\usepackage{epic,eepic,color}
\usepackage{times}
\usepackage{mathptmx}
\usepackage{multirow}
\usepackage{stkernel}

\usepackage{cases}

\usepackage{times}
\usepackage{multirow}

\definecolor{red}{rgb}{1,0,0}
\definecolor{green}{rgb}{0,1,0}
\definecolor{blue}{rgb}{0,0,1}
\definecolor{violet}{rgb}{1,0,1}
\definecolor{cyan}{cmyk}{1,0,0,0}
\definecolor{magenta}{cmyk}{0,1,0,0}
\definecolor{yellow}{cmyk}{0,0,1,0}

\definecolor{white}{rgb}{1,1,1}

\usepackage{jumoline}

\newcommand{\CO}[1]{}

\CO{

\onecolumn
\textwidth=10cm
}

\newcommand{\CommentOut}[1]{}

\begin{document}

\newcommand{\FIG}[3]{
\begin{minipage}[b]{#1cm}
\begin{center}
\includegraphics[width=#1cm]{#2}
{\scriptsize #3}
\end{center}
\end{minipage}
}

\newcommand{\FIGm}[3]{
\begin{minipage}[b]{#1cm}
\begin{center}
\includegraphics[width=#1cm]{#2}\vspace*{-2mm}\\
{\scriptsize #3}
\end{center}
\end{minipage}
}

\newcommand{\FIGR}[3]{
\begin{minipage}[b]{#1cm}
\begin{center}
\includegraphics[angle=-90,clip,width=#1cm]{#2}\vspace*{1mm}
\\
{\scriptsize #3}
\vspace*{1mm}
\end{center}
\end{minipage}
}

\newcommand{\FIGRpng}[5]{
\begin{minipage}[b]{#1cm}
\begin{center}
\includegraphics[bb=0 0 #4 #5, angle=-90,clip,width=#1cm]{#2}\vspace*{1mm}
\\
{\scriptsize #3}
\vspace*{1mm}
\end{center}
\end{minipage}
}

\newcommand{\FIGpng}[5]{
\begin{minipage}[b]{#1cm}
\begin{center}
\includegraphics[bb=0 0 #4 #5, clip, width=#1cm]{#2}\vspace*{-1mm}\\
{\scriptsize #3}
\vspace*{1mm}
\end{center}
\end{minipage}
}

\newcommand{\FIGtpng}[5]{
\begin{minipage}[t]{#1cm}
\begin{center}
\includegraphics[bb=0 0 #4 #5, clip,width=#1cm]{#2}\vspace*{1mm}
\\
{\scriptsize #3}
\vspace*{1mm}
\end{center}
\end{minipage}
}

\newcommand{\FIGRt}[3]{
\begin{minipage}[t]{#1cm}
\begin{center}
\includegraphics[angle=-90,clip,width=#1cm]{#2}\vspace*{1mm}
\\
{\scriptsize #3}
\vspace*{1mm}
\end{center}
\end{minipage}
}

\newcommand{\FIGRm}[3]{
\begin{minipage}[b]{#1cm}
\begin{center}
\includegraphics[angle=-90,clip,width=#1cm]{#2}\vspace*{0mm}
\\
{\scriptsize #3}
\vspace*{1mm}
\end{center}
\end{minipage}
}

\newcommand{\FIGC}[5]{
\begin{minipage}[b]{#1cm}
\begin{center}
\includegraphics[width=#2cm,height=#3cm]{#4}~$\Longrightarrow$\vspace*{0mm}
\\
{\scriptsize #5}
\vspace*{8mm}
\end{center}
\end{minipage}
}

\newcommand{\FIGf}[3]{
\begin{minipage}[b]{#1cm}
\begin{center}
\fbox{\includegraphics[width=#1cm]{#2}}\vspace*{0.5mm}\\
{\scriptsize #3}
\end{center}
\end{minipage}
}

\author{Murase Tomoya ~~~~~~~~~~~~ Tanaka Kanji
\thanks{Our work has been supported in part by 
JSPS KAKENHI 
Grant-in-Aid for Young Scientists (B) 23700229,
and for Scientific Research (C) 26330297 (``The realization of next-generation,
discriminative and succinct SLAM technique: PartSLAM'').}
\thanks{T. Murase and K. Tanaka are with Graduate School of Engineering, University of Fukui, Japan.
{\tt\small tnkknj@u-fukui.ac.jp}}
\vspace*{-5mm}}

\thispagestyle{empty}
\pagestyle{empty}

\newcommand{\figA}{
\begin{figure}[t]
\begin{center}
\FIG{8}{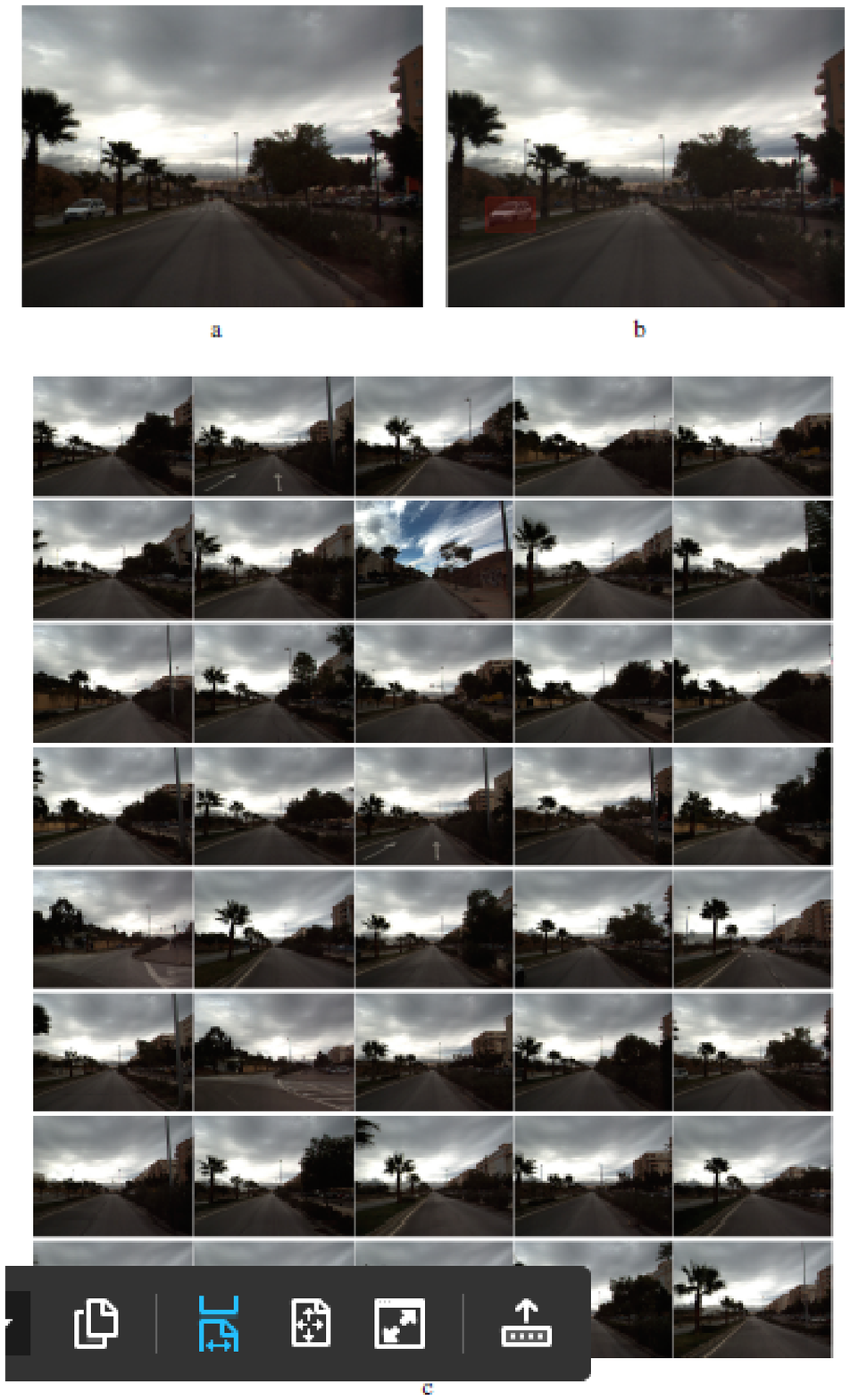}{}
\caption{Image retrieval: (a) query image, (b) ground-truth moving object, and (c) retrieved reference images.}\label{fig:A}
\end{center}
\end{figure}
}

\newcommand{\figB}{
\begin{figure}[t]
\begin{center}
\FIG{8}{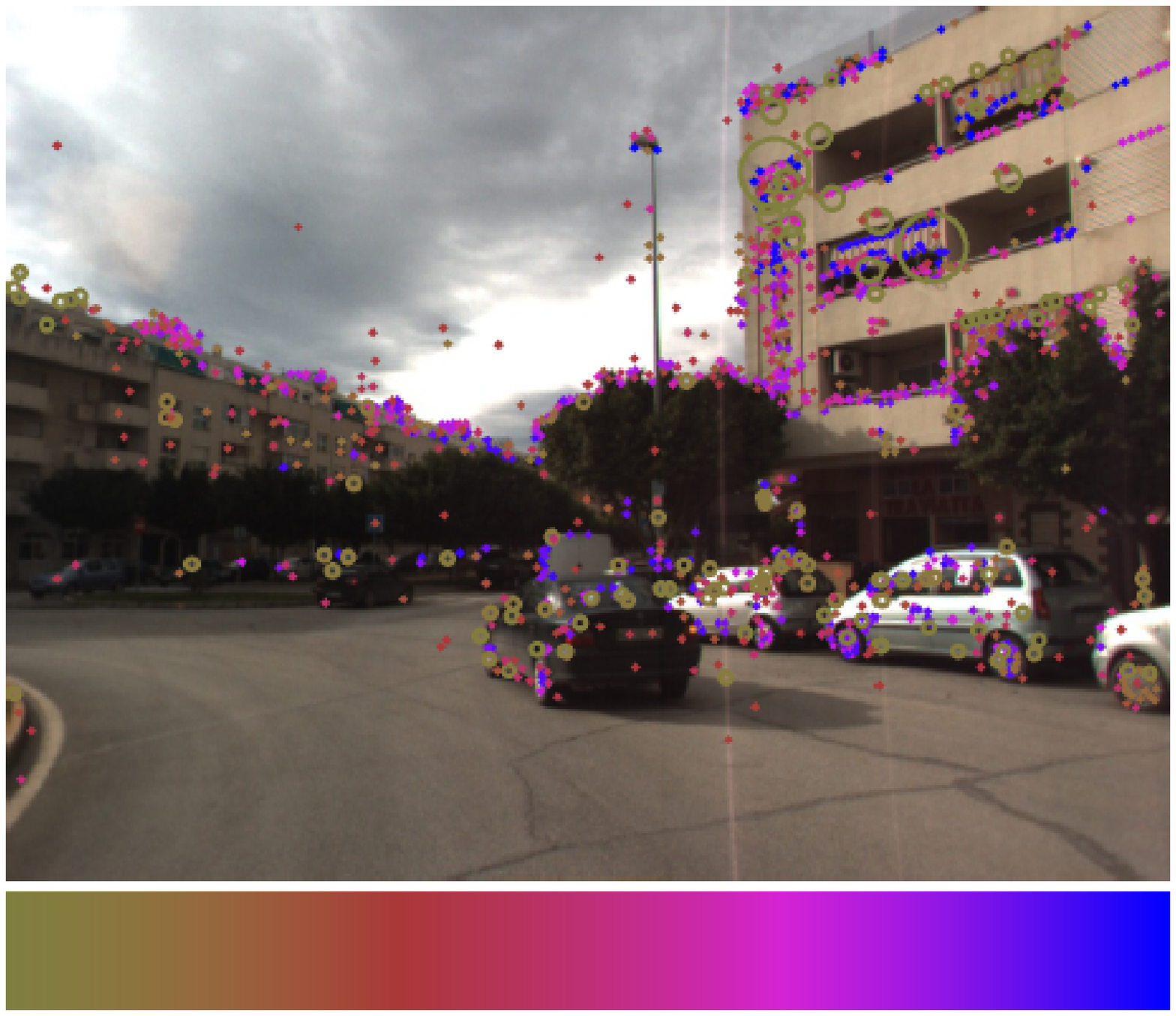}{}
\caption{Compressive change retrieval for moving object detection.
  In our change detection formulation, the robot's current view image is compared against reference images that are retrieved from street-view images using the view image as query. Shown in the figure is an example of change detection result: detected anomaly (small colored points) and anomalyness score (color bar).}\label{fig:B}
\end{center}
\end{figure}
}

\newcommand{\figC}{
\begin{figure*}[t]
\begin{center}
  \FIG{17}{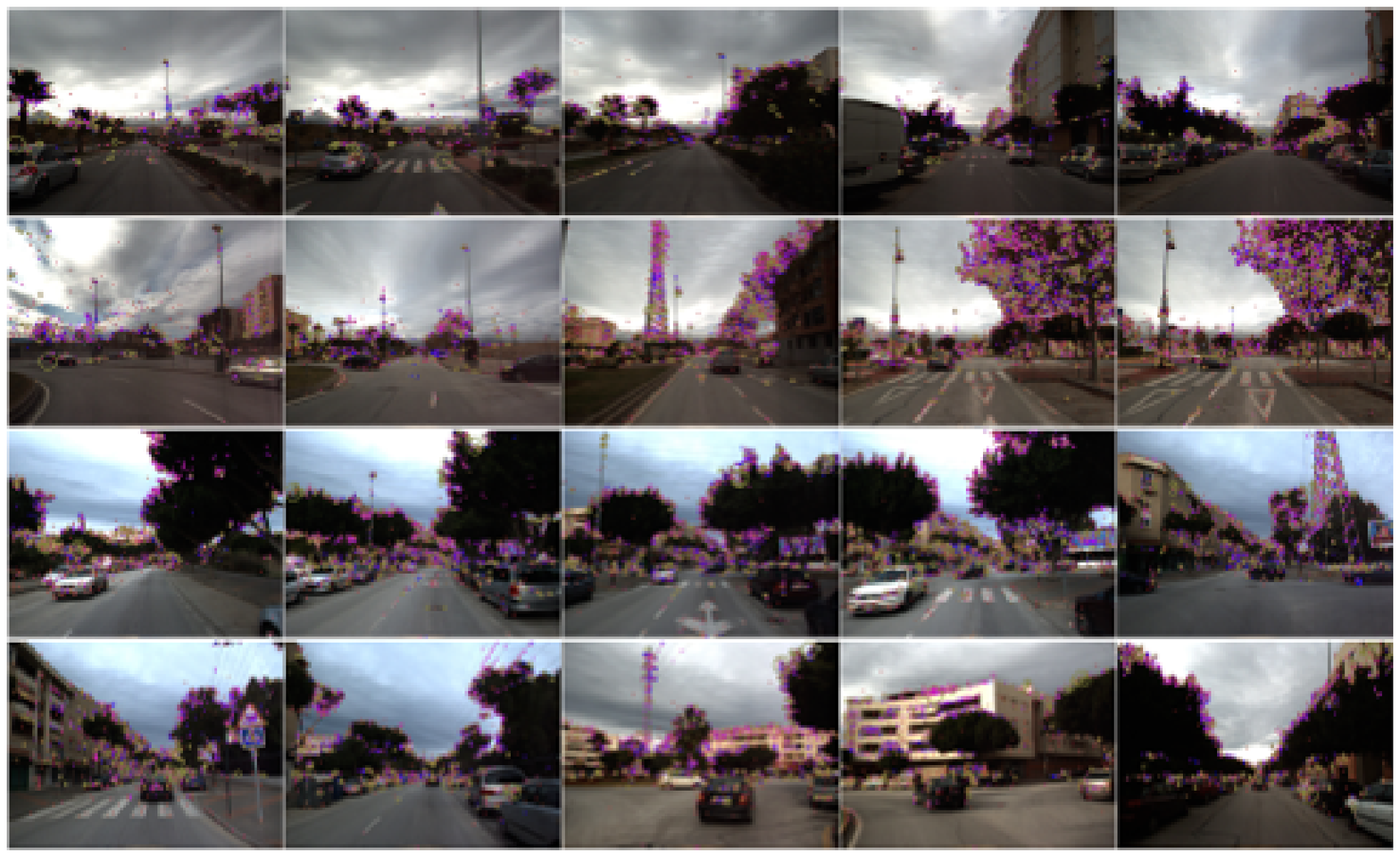}{}
\caption{Results of change detection.}\label{fig:C}
\end{center}
\end{figure*}
}

\newcommand{\figD}{
\begin{figure}[t]
\begin{center}
\FIG{8}{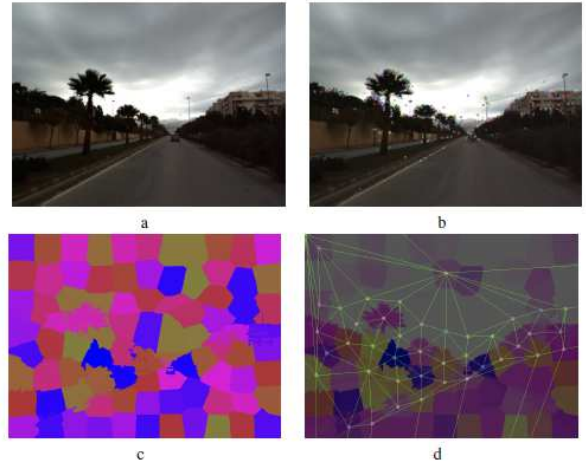}{}
\caption{Image processing.
(a) Input image. (b) SIFT keypoints. (c) Superpixel regions.
  (d) Result of local geometry analysis.
  The points indicate center points of superpixel regions.
  The lines connect geometric neihbor points.
  Only those features inside the visibility region are used.
}\label{fig:D}
\end{center}
\end{figure}
}

\newcommand{\figE}{
\begin{figure}[t]
\begin{center}
  \FIG{8.5}{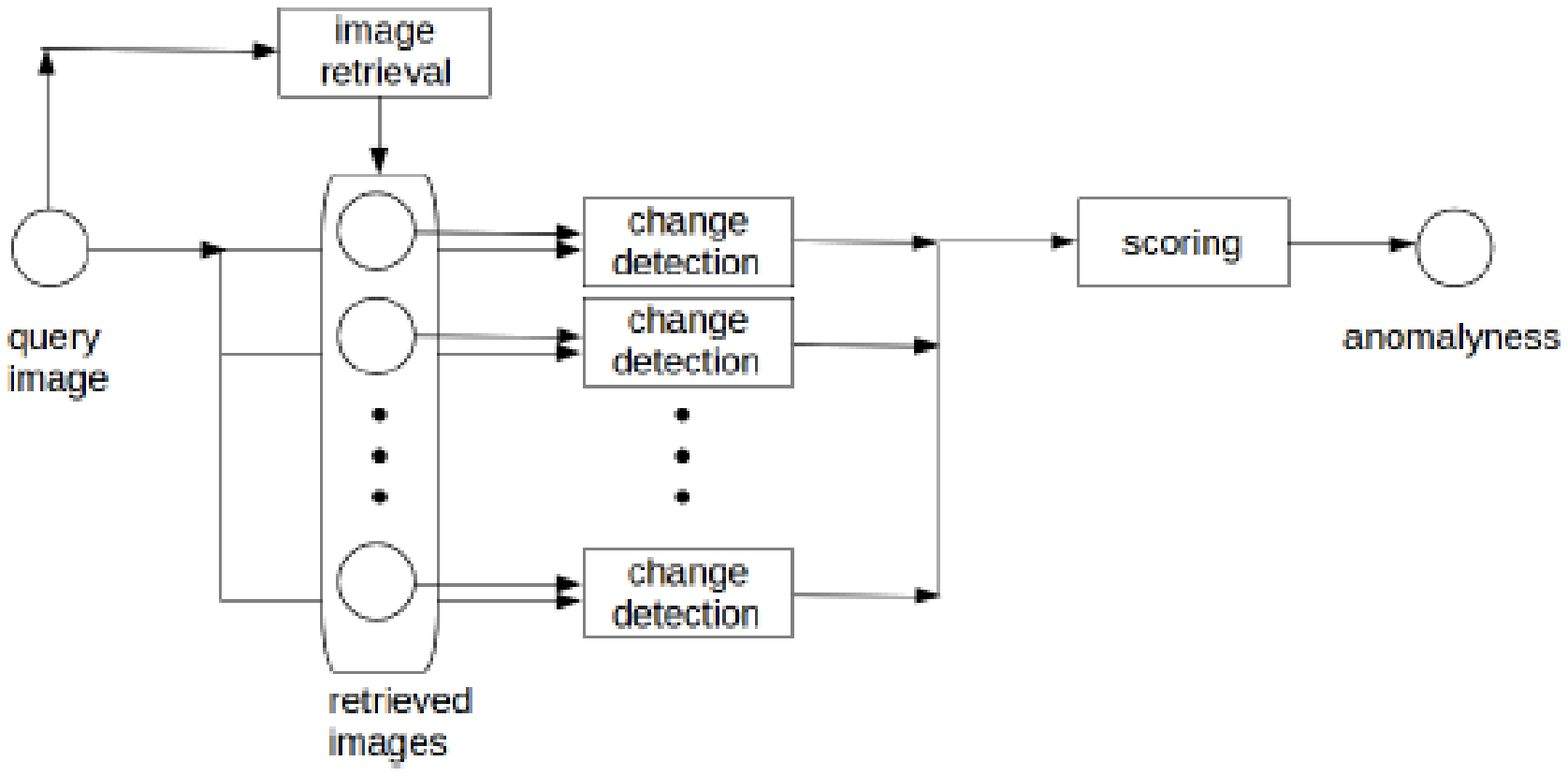}{}
\caption{Main processing.}\label{fig:E}
\vspace*{-7mm}
\end{center}
\end{figure}
}

\newcommand{\figG}{
\begin{figure}[t]
\begin{center}
\FIG{8}{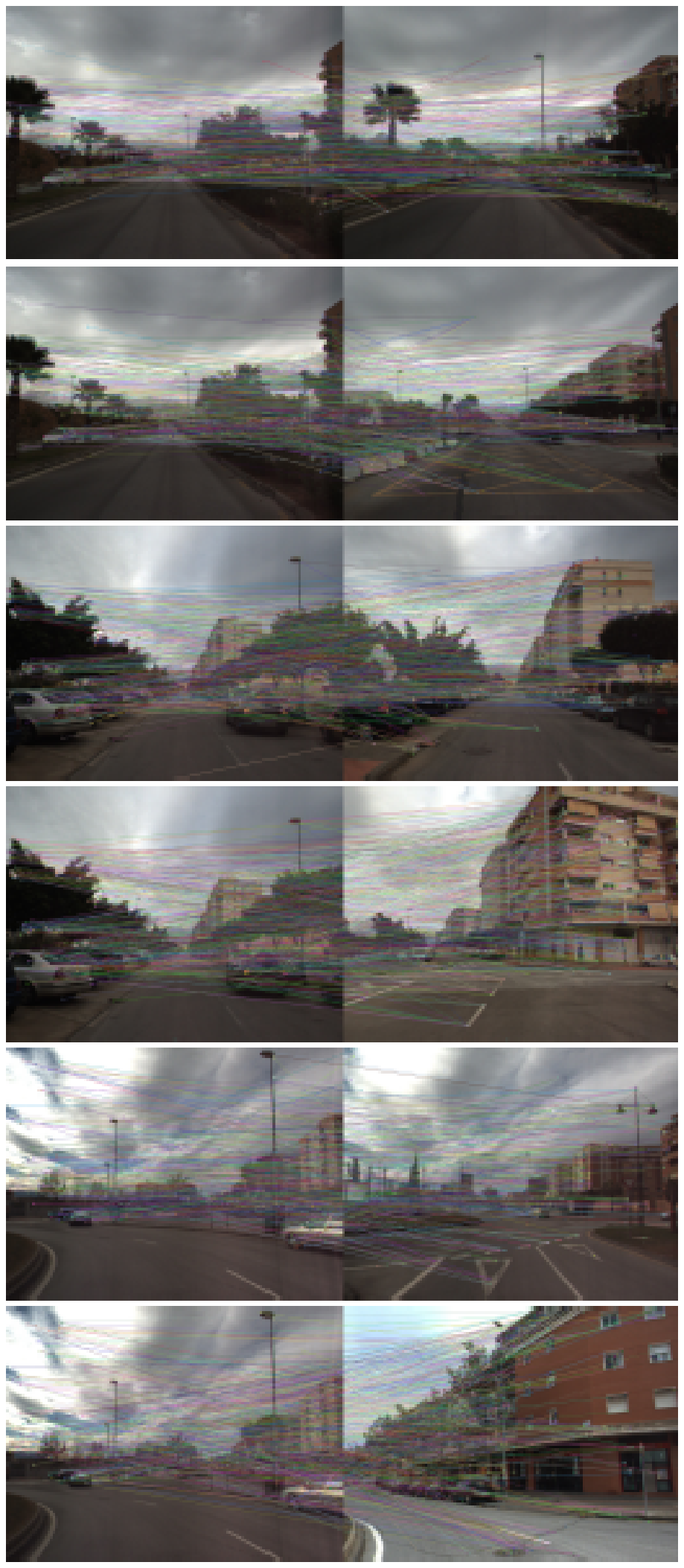}{}
\caption{Matching BoW features.}\label{fig:G}
\end{center}
\end{figure}
}

\newcommand{\tabC}{
\begin{table}[t]
\begin{center}
\caption{Change detection performance in [\%]}\label{tab:C}
\begin{tabular}{|l|l|r|r|r|r|r|r|r|r|r|r|r|r|r|r|} \hline
\#ref & algorithm \hspace*{-2mm}&\hspace*{-2mm} top-1 \hspace*{-2mm}&\hspace*{-2mm} top-2 \hspace*{-2mm}&\hspace*{-2mm} top-5 \hspace*{-2mm}&\hspace*{-2mm} top-10 \hspace*{-2mm}&\hspace*{-2mm} top-20 \hspace*{-2mm}&\hspace*{-2mm} top-50\\\hline \hline 
40 &DM & 17.6 & 20.7 & {\bf 39.2} & {\bf 54.7} & 62.9 & 81.5 \\
   &DM+LG & 17.6 & 20.7 & {\bf 39.2} & {\bf 54.7} & 62.9 & 81.5 \\
   &CCR & {\bf 18.6} & 20.7 & 31.0 & 47.5 & 64.0 & 81.5 \\
   &CCR+LG & 13.5 & {\bf 23.8} & {\bf 39.2} & 51.6 & {\bf 70.2} & {\bf 88.7} \\
   &CCR+LG+VA & 12.4 & 22.7 & 36.1 & 46.4 & 67.1 & 83.6 \\\hline \hline 
20 &DM & 13.5 & 18.6 & 29.9 & 42.3 & 61.9 & 81.5 \\
   &DM+LG & {\bf 20.7} & {\bf 24.8} & {\bf 32.0} & {\bf 49.5} & 65.0 & {\bf 83.6} \\
   &CCR & 13.5 & 21.7 & {\bf 32.0} & 46.4 & {\bf 66.0} & 80.5 \\
   &CCR+LG & 12.4 & 19.6 & 28.9 & 44.4 & 60.9 & 79.4 \\
   &CCR+LG+VA & 15.5 & 22.7 & {\bf 32.0} & 45.4 & 60.9 & 80.5 \\\hline \hline 
10 &DM & {\bf 17.6} & 21.7 & 27.9 & 44.4 & 64.0 & 80.5 \\
   &DM+LG & {\bf 17.6} & {\bf 23.8} & {\bf 32.0} & {\bf 46.4} & 60.9 & 81.5 \\
   &CCR & 10.4 & 18.6 & 31.0 & 45.4 & {\bf 65.0} & {\bf 82.5} \\
   &CCR+LG & 11.4 & 14.5 & 26.9 & 40.3 & 57.8 & 81.5 \\
   &CCR+LG+VA & 12.4 & 15.5 & 24.8 & 38.2 & 55.7 & 80.5 \\\hline \hline 
5 &DM & {\bf 18.6} & {\bf 21.7} & 31.0 & 46.4 & 61.9 & 76.3 \\
   &DM+LG & 12.4 & 19.6 & {\bf 36.1} & {\bf 52.6} & 62.9 & 78.4 \\
   &CCR & 14.5 & 18.6 & 34.1 & 44.4 & {\bf 66.0} & {\bf 81.5} \\
   &CCR+LG & 15.5 & 20.7 & 34.1 & 45.4 & 58.8 & 79.4 \\
   &CCR+LG+VA & 13.5 & 20.7 & 35.1 & 43.3 & 57.8 & 79.4 \\\hline \hline 
1 &DM & 12.4 & 15.5 & 28.9 & 39.2 & 59.8 & 78.4 \\
   &DM+LG & 16.5 & 20.7 & 31.0 & 38.2 & 64.0 & {\bf 81.5} \\
   &CCR & 15.5 & 18.6 & 31.0 & 45.4 & 59.8 & 78.4 \\
   &CCR+LG & {\bf 22.7} & {\bf 22.7} & {\bf 41.3} & {\bf 51.6} & {\bf 66.0} & 80.5 \\
   &CCR+LG+VA & 19.6 & 21.7 & 39.2 & 49.5 & 62.9 & 78.4 \\
  \hline
\end{tabular}\vspace*{1mm}\\
(
DM: direct matching of raw SIFT features.
CCR: compressive change retrieval of BoW.
LG: local geometry.
VA: visibility analysis.
)
\end{center}
\end{table}
}

\newcommand{\tabD}{
\begin{table}[t]
\begin{center}
\caption{Performance gain}\label{tab:D}
\begin{tabular}{|l|l|r|r|r|r|r|r|r|r|r|r|r|r|r|r|} \hline
\#ref & algorithm \hspace*{-2mm}&\hspace*{-2mm} top-1 \hspace*{-2mm}&\hspace*{-2mm} top-2 \hspace*{-2mm}&\hspace*{-2mm} top-5 \hspace*{-2mm}&\hspace*{-2mm} top-10 \hspace*{-2mm}&\hspace*{-2mm} top-20 \hspace*{-2mm}&\hspace*{-2mm} top-50\\\hline \hline 
40 &DM \hspace*{-2mm}&\hspace*{-2mm}  0.0 \hspace*{-2mm}&\hspace*{-2mm}  0.0 \hspace*{-2mm}&\hspace*{-2mm} {\bf  0.0} \hspace*{-2mm}&\hspace*{-2mm} {\bf  0.0} \hspace*{-2mm}&\hspace*{-2mm}  0.0 \hspace*{-2mm}&\hspace*{-2mm}  0.0 \\
   &DM+LG \hspace*{-2mm}&\hspace*{-2mm}  0.0 \hspace*{-2mm}&\hspace*{-2mm}  0.0 \hspace*{-2mm}&\hspace*{-2mm} {\bf  0.0} \hspace*{-2mm}&\hspace*{-2mm} {\bf  0.0} \hspace*{-2mm}&\hspace*{-2mm}  0.0 \hspace*{-2mm}&\hspace*{-2mm}  0.0 \\
   &CCR \hspace*{-2mm}&\hspace*{-2mm} {\bf +1.1} \hspace*{-2mm}&\hspace*{-2mm}  0.0 \hspace*{-2mm}&\hspace*{-2mm} -8.3 \hspace*{-2mm}&\hspace*{-2mm} -7.3 \hspace*{-2mm}&\hspace*{-2mm} +1.1 \hspace*{-2mm}&\hspace*{-2mm}  0.0 \\
   &CCR+LG \hspace*{-2mm}&\hspace*{-2mm} -4.2 \hspace*{-2mm}&\hspace*{-2mm} {\bf +3.1} \hspace*{-2mm}&\hspace*{-2mm} {\bf  0.0} \hspace*{-2mm}&\hspace*{-2mm} -3.1 \hspace*{-2mm}&\hspace*{-2mm} {\bf +7.3} \hspace*{-2mm}&\hspace*{-2mm} {\bf +7.3} \\
   &CCR+LG+VA \hspace*{-2mm}&\hspace*{-2mm} -5.2 \hspace*{-2mm}&\hspace*{-2mm} +2.1 \hspace*{-2mm}&\hspace*{-2mm} -3.1 \hspace*{-2mm}&\hspace*{-2mm} -8.3 \hspace*{-2mm}&\hspace*{-2mm} +4.2 \hspace*{-2mm}&\hspace*{-2mm} +2.1 \\\hline \hline 
20 &DM \hspace*{-2mm}&\hspace*{-2mm}  0.0 \hspace*{-2mm}&\hspace*{-2mm}  0.0 \hspace*{-2mm}&\hspace*{-2mm}  0.0 \hspace*{-2mm}&\hspace*{-2mm}  0.0 \hspace*{-2mm}&\hspace*{-2mm}  0.0 \hspace*{-2mm}&\hspace*{-2mm}  0.0 \\
   &DM+LG \hspace*{-2mm}&\hspace*{-2mm} {\bf +7.3} \hspace*{-2mm}&\hspace*{-2mm} {\bf +6.2} \hspace*{-2mm}&\hspace*{-2mm} {\bf +2.1} \hspace*{-2mm}&\hspace*{-2mm} {\bf +7.3} \hspace*{-2mm}&\hspace*{-2mm} +3.1 \hspace*{-2mm}&\hspace*{-2mm} {\bf +2.1} \\
   &CCR \hspace*{-2mm}&\hspace*{-2mm}  0.0 \hspace*{-2mm}&\hspace*{-2mm} +3.1 \hspace*{-2mm}&\hspace*{-2mm} {\bf +2.1} \hspace*{-2mm}&\hspace*{-2mm} +4.2 \hspace*{-2mm}&\hspace*{-2mm} {\bf +4.2} \hspace*{-2mm}&\hspace*{-2mm} -1.1 \\
   &CCR+LG \hspace*{-2mm}&\hspace*{-2mm} -1.1 \hspace*{-2mm}&\hspace*{-2mm} +1.1 \hspace*{-2mm}&\hspace*{-2mm} -1.1 \hspace*{-2mm}&\hspace*{-2mm} +2.1 \hspace*{-2mm}&\hspace*{-2mm} -1.1 \hspace*{-2mm}&\hspace*{-2mm} -2.1 \\
   &CCR+LG+VA \hspace*{-2mm}&\hspace*{-2mm} +2.1 \hspace*{-2mm}&\hspace*{-2mm} +4.2 \hspace*{-2mm}&\hspace*{-2mm} {\bf +2.1} \hspace*{-2mm}&\hspace*{-2mm} +3.1 \hspace*{-2mm}&\hspace*{-2mm} -1.1 \hspace*{-2mm}&\hspace*{-2mm} -1.1 \\\hline \hline 
10 &DM \hspace*{-2mm}&\hspace*{-2mm} {\bf  0.0} \hspace*{-2mm}&\hspace*{-2mm}  0.0 \hspace*{-2mm}&\hspace*{-2mm}  0.0 \hspace*{-2mm}&\hspace*{-2mm}  0.0 \hspace*{-2mm}&\hspace*{-2mm}  0.0 \hspace*{-2mm}&\hspace*{-2mm}  0.0 \\
   &DM+LG \hspace*{-2mm}&\hspace*{-2mm} {\bf  0.0} \hspace*{-2mm}&\hspace*{-2mm} {\bf +2.1} \hspace*{-2mm}&\hspace*{-2mm} {\bf +4.2} \hspace*{-2mm}&\hspace*{-2mm} {\bf +2.1} \hspace*{-2mm}&\hspace*{-2mm} -3.1 \hspace*{-2mm}&\hspace*{-2mm} +1.1 \\
   &CCR \hspace*{-2mm}&\hspace*{-2mm} -7.3 \hspace*{-2mm}&\hspace*{-2mm} -3.1 \hspace*{-2mm}&\hspace*{-2mm} +3.1 \hspace*{-2mm}&\hspace*{-2mm} +1.1 \hspace*{-2mm}&\hspace*{-2mm} {\bf +1.1} \hspace*{-2mm}&\hspace*{-2mm} {\bf +2.1} \\
   &CCR+LG \hspace*{-2mm}&\hspace*{-2mm} -6.2 \hspace*{-2mm}&\hspace*{-2mm} -7.3 \hspace*{-2mm}&\hspace*{-2mm} -1.1 \hspace*{-2mm}&\hspace*{-2mm} -4.2 \hspace*{-2mm}&\hspace*{-2mm} -6.2 \hspace*{-2mm}&\hspace*{-2mm} +1.1 \\
   &CCR+LG+VA \hspace*{-2mm}&\hspace*{-2mm} -5.2 \hspace*{-2mm}&\hspace*{-2mm} -6.2 \hspace*{-2mm}&\hspace*{-2mm} -3.1 \hspace*{-2mm}&\hspace*{-2mm} -6.2 \hspace*{-2mm}&\hspace*{-2mm} -8.3 \hspace*{-2mm}&\hspace*{-2mm}  0.0 \\\hline \hline 
5 &DM \hspace*{-2mm}&\hspace*{-2mm} {\bf  0.0} \hspace*{-2mm}&\hspace*{-2mm} {\bf  0.0} \hspace*{-2mm}&\hspace*{-2mm}  0.0 \hspace*{-2mm}&\hspace*{-2mm}  0.0 \hspace*{-2mm}&\hspace*{-2mm}  0.0 \hspace*{-2mm}&\hspace*{-2mm}  0.0 \\
   &DM+LG \hspace*{-2mm}&\hspace*{-2mm} -6.2 \hspace*{-2mm}&\hspace*{-2mm} -2.1 \hspace*{-2mm}&\hspace*{-2mm} {\bf +5.2} \hspace*{-2mm}&\hspace*{-2mm} {\bf +6.2} \hspace*{-2mm}&\hspace*{-2mm} +1.1 \hspace*{-2mm}&\hspace*{-2mm} +2.1 \\
   &CCR \hspace*{-2mm}&\hspace*{-2mm} -4.2 \hspace*{-2mm}&\hspace*{-2mm} -3.1 \hspace*{-2mm}&\hspace*{-2mm} +3.1 \hspace*{-2mm}&\hspace*{-2mm} -2.1 \hspace*{-2mm}&\hspace*{-2mm} {\bf +4.2} \hspace*{-2mm}&\hspace*{-2mm} {\bf +5.2} \\
   &CCR+LG \hspace*{-2mm}&\hspace*{-2mm} -3.1 \hspace*{-2mm}&\hspace*{-2mm} -1.1 \hspace*{-2mm}&\hspace*{-2mm} +3.1 \hspace*{-2mm}&\hspace*{-2mm} -1.1 \hspace*{-2mm}&\hspace*{-2mm} -3.1 \hspace*{-2mm}&\hspace*{-2mm} +3.1 \\
   &CCR+LG+VA \hspace*{-2mm}&\hspace*{-2mm} -5.2 \hspace*{-2mm}&\hspace*{-2mm} -1.1 \hspace*{-2mm}&\hspace*{-2mm} +4.2 \hspace*{-2mm}&\hspace*{-2mm} -3.1 \hspace*{-2mm}&\hspace*{-2mm} -4.2 \hspace*{-2mm}&\hspace*{-2mm} +3.1 \\\hline \hline 
1 &DM \hspace*{-2mm}&\hspace*{-2mm}  0.0 \hspace*{-2mm}&\hspace*{-2mm}  0.0 \hspace*{-2mm}&\hspace*{-2mm}  0.0 \hspace*{-2mm}&\hspace*{-2mm}  0.0 \hspace*{-2mm}&\hspace*{-2mm}  0.0 \hspace*{-2mm}&\hspace*{-2mm}  0.0 \\
   &DM+LG \hspace*{-2mm}&\hspace*{-2mm} +4.2 \hspace*{-2mm}&\hspace*{-2mm} +5.2 \hspace*{-2mm}&\hspace*{-2mm} +2.1 \hspace*{-2mm}&\hspace*{-2mm} -1.1 \hspace*{-2mm}&\hspace*{-2mm} +4.2 \hspace*{-2mm}&\hspace*{-2mm} {\bf +3.1} \\
   &CCR \hspace*{-2mm}&\hspace*{-2mm} +3.1 \hspace*{-2mm}&\hspace*{-2mm} +3.1 \hspace*{-2mm}&\hspace*{-2mm} +2.1 \hspace*{-2mm}&\hspace*{-2mm} +6.2 \hspace*{-2mm}&\hspace*{-2mm}  0.0 \hspace*{-2mm}&\hspace*{-2mm}  0.0 \\
   &CCR+LG \hspace*{-2mm}&\hspace*{-2mm} {\bf +10.4} \hspace*{-2mm}&\hspace*{-2mm} {\bf +7.3} \hspace*{-2mm}&\hspace*{-2mm} {\bf +12.4} \hspace*{-2mm}&\hspace*{-2mm} {\bf +12.4} \hspace*{-2mm}&\hspace*{-2mm} {\bf +6.2} \hspace*{-2mm}&\hspace*{-2mm} +2.1 \\
   &CCR+LG+VA \hspace*{-2mm}&\hspace*{-2mm} +7.3 \hspace*{-2mm}&\hspace*{-2mm} +6.2 \hspace*{-2mm}&\hspace*{-2mm} +10.4 \hspace*{-2mm}&\hspace*{-2mm} +10.4 \hspace*{-2mm}&\hspace*{-2mm} +3.1 \hspace*{-2mm}&\hspace*{-2mm}  0.0 \\
  \hline
\end{tabular}
\end{center}
\end{table}
}

\title{\LARGE \bf
Compressive Change Retrieval for Moving Object Detection
}

\maketitle

\begin{abstract}
Change detection, or anomaly detection, from street-view images acquired by an autonomous robot at multiple different times, is a major problem in robotic mapping and autonomous driving. Formulation as an image comparison task, which operates on a given pair of query and reference images is common to many existing approaches to this problem. Unfortunately, providing relevant reference images is not straightforward. In this paper, we propose a novel formulation for change detection, termed compressive change retrieval, which can operate on a query image and similar reference images retrieved from the web. Compared to previous formulations, there are two sources of difficulty. First, the retrieved reference images may frequently contain non-relevant reference images, because even state-of-the-art place-recognition techniques suffer from retrieval noise. Second, image comparison needs to be conducted in a compressed domain to minimize the storage cost of large collections of street-view images. To address the above issues, we also present a practical change detection algorithm that uses compressed bag-of-words (BoW) image representation as a scalable solution. The results of experiments conducted on a practical change detection task, ``moving object detection (MOD)," using the publicly available Malaga dataset validate the effectiveness of the proposed approach.
\end{abstract}

\section{Introduction}

Change detection, or anomaly detection, from street-view images acquired by an autonomous robot at multiple different times,
is a major problem for robotic mapping and autonomous driving. Given a visual image of the robot's surroundings as a query, the goal of change detection is to search over a database or a collection of previously acquired images to identify regions that correspond to environment changes (e.g., appearance of new objects), which comprise the ``change mask'' \cite{ccr1}. 
A key issue is that the change mask should not contain unimportant or nuisance forms of change, such as those induced by difference in views and sensor noises. In this sense, change detection is similar in its objectives to anomaly detection \cite{ccr2}, 
in which the goal is to detect anomalies that are interesting to the observer.

The problem of change detection from street-view images has drawn much research attention over the past decade \cite{ccr3,ccr4,ccr5}, and has resulted in the production of many interesting and effective algorithms. The use of 3D line segments for change detection, in which line segments are matched between multiple views, 3D line segments reconstructed, and 3D images then compared to detect changes, is proposed in \cite{ccr3}.
Change detection from Google StreetView image panoramas acquired by moving vehicles, in which a coarse 3D geometry of the scene is recovered and then registered with previously acquired reference images of the location, and further semantic content of current and previous views are exploited to gather additional evidence about the change hypothesis, is proposed in 
\cite{ccr4}.
``City-scale" change detection from 3D city models and panoramic images captured by a car driving around the city, in which geometric changes are detected by comparing images and 3D models, under inaccuracies in input geometry, errors in the image's GPS data, as well as limited amount of information owing to sparse imagery, is proposed in \cite{ccr5}.

\figB

Formulation as an {\it image comparison} task, which operates on a given pair of query and reference images, is common to the majority of approaches such as these. Unfortunately, providing relevant reference images is non-trivial and the main topic of ongoing place recognition studies. To date, most state-of-the-art systems simply assume relevant reference images are given, or rely on the availability of a reference image's viewpoint (e.g., GPS), which limits their application scenarios.

In this paper,
we propose a novel formulation for change detection,
termed {\it compressive change retrieval},
which does not require a relevant reference image,
but instead can operate on a query image and similar reference images retrieved from the web (Fig. \ref{fig:B}).
This study is motivated by recent progress in large-scale image retrieval and publicly available street-view images,
thanks to which it is possible to obtain a collection of similar street-view images (i.e., candidate reference images) to a given query image.
Compared to previous formulations, there are two sources of difficulty.
First, the retrieved reference images may frequently contain non-relevant reference images,
because even state-of-the-art image retrieval techniques suffer from retrieval noises.
Second, image comparison needs to be conducted in a compressed domain to minimize the storage cost of large collections of street-view images.
To address the above issues, we present a practical change detection algorithm that uses compressed bag-of-words (BoW) image representation as a scalable solution. The following contributions are made in this paper: (1) We reformulate the change detection task as compressive change retrieval and thereby extend change detection to the case of multiple noisy reference images. (2) We present a practical change detection algorithm for compressed BoW image representation, spatial analysis, and occlusion reasoning. (3) We implement the change detection framework on a practical application of ``moving object detection (MOD)" from street-view images. Experimental results using the publicly available Malaga dataset validate the effectiveness of the proposed approach.

Our approach is orthognal to most of the existing approaches to MOD.
The experimental scenario that we consider for change detection is a practical application of MOD,
where the goal is to detect moving objects (e.g., cars) in a single urban image by comparing the query image with similar retrieved images.
Existing solutions to moving object detection
can be broadly categorized into
those using motion cues (e.g., motion segmentation, and moving camera background subtraction)
and those based on prior knowledge (e.g., pre-trained object detector, and change detection).
The types of prior knowledge used in the latter category
can be divided into
prior on foreground (i.e., labeled object examples)
and
prior on background (i.e., image acquired at different times).
All of the above approaches are far from perfect,
and suffer from
image processing noise,
offline costs for acquiring prior,
or both.
Our proposed approach can be viewed as
a prior based approach with a novel low cost prior (i.e., random street-view images),
which has not been sufficiently explored in the literature
and is the main contribution of our study.

\figA

\section{Approach}\label{sec:appr}

For clarity of presentation, we first describe the baseline approach, which is the base for our approach and is a performance comparison benchmark in the experiments described in section \ref{sec:exp}. Then, we present a solution to the change retrieval task as an extension of the baseline system.

\figE

\subsection{Baseline System}

The main process consists of three steps (Fig.\ref{fig:E}): (1) image retrieval, (2) change proposal, and (3) decision. 
In the first step,
the aim is to obtain a set of candidate reference images
by retrieving the collection of street-view images. 

For image retrieval, any visual features (e.g., SIFT, SURF) can be used but we chose DCNN features from
Alexnet \cite{alexnet}
because of their excellent performance.
Fig. \ref{fig:A}
shows an example of the image retrieval process.
As can be seen,
relevant images are successfully retrieved
thanks to the DCNN features.
However,
it can also be seen that
there is a non-negligible number of non-relevant images,
due to the retrieval noise.

In the second step, the aim is to compare visual local features between the query image and each reference image of interest to evaluate the probability of individual features in the query image corresponding to changes. To this end, we employed 128-dim SIFT descriptors at interest keypoints as features, and computed the dissimilarity in terms of L2 norm as anomalyness. 

The third step integrates the results of scene comparisons for all the reference images and, based on the result, we compute the anomalyness of each feature in the query image. The basic idea is to compute the minimum of the dissimilarity (i.e., L2 norm) over all the reference images and view it as the anomalyness of the given feature:
\begin{equation}
  A(f^q) = \min_{r\in R} \min_{f^r\in r} | f^q - f^r |_2,
\end{equation}
where $R$ is the set of reference images, and $f^q$ and $f^r$ are features in the query and a reference image $r$.

\subsection{Bag-of-Words Extension}

The basic idea of underlying bag-of-words image representation is
translation of the visual local features in a given image
to an unordered collection of visual words.
In preprocessing,
we prepare a fine visual vocabulary
consisting of 1M exemplar visual features
each of which corresponds to a different visual word.
To translate a given feature in a reference image,
we search over the vocabulary to find the feature's nearest neighbor exemplar,
and assign the exempler's ID as the feature's visual word.
The result is an unordered collection of visual words,
termed bag-of-words.
All the features in all the street-view images
are stored in the inverted index of visual words
and retrieved therefrom.

Our strategy for dissmilarity evaluation
is an instance of 
asymmetric distance computation (ADC) \cite{iros16a},
which only encodes the local features of the reference images,
not the local feature in the query image.
This is in contrast to symmetric distance
computation (SDC) employed by typical BoW systems,
which encodes both query and database features.
Our ADC-based method
runs kNN search over the set intersection between the reference images' features and the vocabulary's features using the given query feature.
Then, it computes the L2 distance between the query feature
and the nearest neighbor feature as the anomalyness of the query feature.

\figD

\subsection{Local Geometry}\label{sec:lg}

Alignment is a standard preprocessing step in change detection \cite{ccr1} that aligns local features between query and reference images. This preprocessing enables change detection algorithms to compare only those features that are spatially near to each other, and reduces incidences of similar but spatially distant objects not being detected. A naive strategy is to perform one-to-one registration for every pair of query and reference images. In our case, it is unfortunately impractical to run the alignment step for all the pairs of query and reference images, as the collection of street-view images is large. Instead, we here describe an efficient strategy for the alignment that can operate on the efficient inverted index.

In preprocessing,
we analyze the LG of local feature positions
and store the result in the inverted index (Fig.\ref{fig:D}).
To this end,
we adapt 
triangulation-based analysis of LG,
which was originally proposed
for a different application, specifically, ``scalable logo recognition'', in \cite{ccr7}. 
We capture feature geometry using Delauney triangulation
on local feature positions.
Because a Delauney graph is quite sensitive to the position's noise,
we employ superpixel image segmentation in \cite{achanta2010slic} 
and then each superpixel's center position
acts as the position of the features that belong to the superpixel.
To this end,
we need to obtain and store
two kinds of information:
(1) the IDs of the superpixels that are adjacent to each superpixel,
and
(2) the ID of superpixel to which each feature belongs.

The alignment step
first establishes correspondence
between each feature in the query image
and visual words in the reference image
to find {\it strong matches}
$\{ 1, \cdots, K \}$
that satisfy the SIFT ratio condition in \cite{lowe2004distinctive}:
\begin{equation}
  dist[1] < \cdots < dist[K] < 0.8 * dist[K+1],
\end{equation}
where $d[k]$ is the L2 distance between the query feature of interest
and the $k$-th nearest neighbor match.
If strong matches are found (i.e., $K\ge 1$),
each query feature is
compared with those reference features that belong to a selected subset of superpixels
(rather than all the features in the reference image),
which
are adjacent to those superpixels
that have at least one strong match.

\subsection{Visibility Analysis}\label{sec:va}

We also estimate image regions commonly visible from both query and reference images, which enables change detection algorithms to compare only those features that belong to the commonly visible regions. In this study, the visible regions were simply modeled as a pair of bounding boxes 
$[x_{min}^i, x_{max}^i]$
$\times$
$[y_{min}^i, y_{max}^i]$
$(i \in \{ q, r \})$
in query ($q$)  and  reference  ($r$)  images. To determine the  bounding  box  for either image, we first sort the matched visual features in the image in ascending order of $x,y$-coordinate, and define
$0.1 * |M^i|$-th
and 
$0.9 * |M^i|$-th
elements
($\delta=0.1$)
in the sorted list as the 
$x,y$-locations 
 of the vertexes of the bounding box. Then, we further enlarge the width and height of the bounding box by a scaling factor of 
$1/(1-\delta)$
to increase robustness. In change detection, we use only those features that belong to superpixels whose center positions lie within the bounding boxes.

\section{Experiments}\label{sec:exp}

We verified our method on a large collection of urban images and compared the results obtained to the baseline change detection method. The experimental scenario considered for change detection was a practical application of  ``moving  object  detection  (MOD)," in which  the  goal is to detect moving objects (e.g., cars) in a given query image by comparing the query image with the retrieved similar images. Such detection is a practical scenario because existing solutions for MOD from motion segmentation, such as moving camera background subtraction, are far from perfect, and even learning-based techniques are ill-posed owing to visual diversity of the appearance of an object. This is also a challenging scenario because the experimental environment contains many similar places and perceptual aliasing makes place recognition a difficult task, and yields retrieval noise. Furthermore, MOD solely from an object's appearance is difficult because there are many non-moving objects (e.g., parking cars) with quite similar models.

Fig. \ref{fig:G} shows the results of 1-NN BoW matching for each query feature over features in a reference image. As can be seen, even 1-NN matches contain both relevant and irrelevant features, which make our anomaly detection task a challenging one. Fig. \ref{fig:C} shows examples of change detection. It can be seen that the proposed method is successful even when there are many similar non-changed cars within one query image.

\subsection{Settings}

For the evaluation, we used image sequences from the Malaga dataset \cite{blanco2009collection}. 
The Malaga dataset contains GPS data,
two cameras facing forward in the direction of vehicle motion,
as well as LIDAR data.
We use the left eye's images
with a resolution of $1024\times 768$
for our change detection task.
Although our application scenario is monoclular visual recognition,
we employed
the stereo image sequence to collect
ground-truth viewpoint information for place recognition and change detection,
where
stereo SLAM with visual odometry 
with loop closure constraints
is used for estimating trajectry.

We considered a practical place recognition scenario, called loop closing \cite{ccr8}, 
derived from the field of visual SLAM,
in which a robot traverses a loop-like trajectory and then returns to the previously explored
location. More formally, the relevant image pair is defined by a database image that satisfies two conditions:
1) Its viewpoint from each query's viewpoint is nearer than other candidates.
2) Its distance traveled along the robot's trajectory is sufficiently larger (200 frames) than that of the query image.
Owning to condition 2,
a relevant pair of images become dissimilar to each other, which makes our place recognition
and change detection tasks challenging.
To this end,
we used datasets \#5, \#6, \#7, \#8, and \#10 from the Malaga dataset,
because they have ``loop-closing'' situations,
each of which consists of 
4816, 
4618,
2121,
10026,
and 17310 images.

We implemented the proposed method in C++.
In accordance with \cite{ccr4}, we used SIFT \cite{lowe2004distinctive} as visual features for image comparison and change detection.
For the superpixel segmentation, we used SLIC superpixel code \cite{achanta2010slic}  with parameters
$nr=100$ and $nc=50$.
In the above settings,
the number of SIFT features extracted per image
was in the range 0-5551.
For bag-of-words translation, we used a visual vocabulary with 1M words.
The default number of candidate reference images per query was set at 40.
The techniques described in subsections \ref{sec:lg} and \ref{sec:va} are respectively termed ``local geometory (LG)" and ``visibility analysis (VA)" below.

To evaluate different methods under comparison, we used a ranking-based performance measure,
for which a smaller value indicates better performance.
The Ranking is defined as the rank of the feature that belongs to the ground-truth location of changed objects,
in a list of features sorted in descending order of the anomalyness.

We selected 110 images at random as query images for the experiments, such that for each moving object that the camera encountered  one or a few query images appeared. We annotated the ground-truth location of changed objects with a form of bounding box in the query image. When there were multiple features in the bounding box, we selected and used the feature with the highest anomalyness score for computation of the Ranking.

\figG

\figC

\tabC
\tabD

\subsection{Results}

We also compared the results obtained by the proposed change detection algorithm with those from the baseline algorithm
using non-quantized SIFT features.
For fair comparison, 
both the proposed and the baseline algorithms
employ the techniques
described in \ref{sec:appr}.
Tables \ref{tab:C} ``CCR'' and ``baseline''
show the
top-$X\%$
($X\in \{1, 3, 5\}$)
detection performance
for the proposed method (CCR)
and the baseline method (baseline).
It can be seen that
the proposed method is comparable
to the baseline method
despite the fact that
the propopsed method is
based on vector-quantized representation of images,
i.e., bag-of-words,
which is significantly more compact.

We set the default number of candidate reference images per query as 40.
This means that for change detection,
each query image was compared against 40 different similar reference images retrieved from the database.
We tested several different values for the number of reference images.
Tables \ref{tab:D}
shows the results of top-3\% NR performance for the values,
40, 20, 10, and 5.
 As expected, the performance of the proposed CCR method was increasingly better as the reference image set became larger and more informative.

We also compared
the cases with and without the 
``Local Geomtry (LG)'' and ``Visibility Analysis (VA)''
techniques described in \ref{sec:lg} and \ref{sec:va}.
In Table \ref{tab:D},
``CCR," ``CCR+LG," and ``CCR+LG+VA" indicate the results with and without the LG and VA techniques. It can be seen that, compared to the case of the proposed algorithm, in both cases, performance fell. In particular, the performance decrease is significant in the case of CCR without LG. Thus, it is clear that the proposed LG and VA techniques both effectively improve change detection performance in the proposed CCR framework.

\section{Conclusions}

In this paper, we addressed the problem of compressive change retrieval for moving object detection (MOD). Unlike existing algorithms for change detection that operate on given reference images, we presented a practical solution that can operate on large collections of unorganized street-view images. Three contributions were made in this paper: (1) We reformulated the change detection task as compressive change retrieval and extended the previous change detection to the case of multiple noisy reference images. (2) We presented a practical change detection algorithm for compressed BoW image representation, spatial analysis, and occlusion reasoning. (3) We implemented the change detection framework on a practical application of MOD. Experimental results using the publicly available Malaga dataset validate the effectiveness of the proposed approach. Our future work will focus on an integrated MOD, in which the proposed detection algorithm will be combined with other recognition algorithms based on motion cues, background subtraction, and visual learning, in order to further improve its change detection performance.

\bibliographystyle{IEEEtran}
\bibliography{ccr}

\end{document}